\documentclass[11pt,a4paper]{article}
\usepackage[hyphens,spaces,obeyspaces]{url}

\usepackage[hyperref]{acl2020}
\usepackage{times}
\usepackage{graphicx}
\usepackage{latexsym}

\usepackage{color}
\usepackage{hhline}
\usepackage{multirow}
\usepackage{makecell}
\usepackage{tabu}
\usepackage{booktabs}
\usepackage{xcolor}

\usepackage{times}
\usepackage{latexsym}

\usepackage{graphicx}
\usepackage{color}
\usepackage{listing}
\usepackage{helvet}  
\usepackage{courier}  
\usepackage{multirow}
\usepackage{amsthm}
\usepackage{amssymb}
\usepackage{amsmath}
\usepackage{color}
\usepackage{comment}
\usepackage{MnSymbol}
\usepackage{makecell}
\usepackage{arydshln}
\usepackage[shortlabels]{enumitem}

\usepackage{microtype}

\urldef{\giturl}\url{https://github.com/WasifurRahman/BERT_multimodal_transformer}

\aclfinalcopy 
\def\aclpaperid{1977} 

\title{Integrating Multimodal Information in Large Pretrained Transformers}

\author{
Wasifur Rahman\textsuperscript{1}, Md. Kamrul Hasan\textsuperscript{1*}, Sangwu Lee\textsuperscript{1*},  Amir Zadeh\textsuperscript{2},\\ \textbf{ Chengfeng Mao\textsuperscript{2}, Louis-Philippe Morency\textsuperscript{2}, Ehsan Hoque\textsuperscript{1}} \\
1 - Department of Computer Science, University of Rochester, USA\\
2 - Language Technologies Institute, SCS, CMU, USA\\
\tt echowdh2@ur.rochester.edu, mhasan8@cs.rochester.edu,\\
\tt slee232@u.rochester.edu,abagherz@cs.cmu.edu,\\
\tt chengfem@andrew.cmu.edu,morency@cs.cmu.edu,\\
\tt mehoque@cs.rochester.edu}

\date{}
\usepackage{color}

\date{}
\makeatletter
\def\blfootnote{\xdef\@thefnmark{}\@footnotetext}
\makeatother

\begin{document}
\maketitle
\begin{abstract}

Recent Transformer-based contextual word representations, including BERT and XLNet, have shown state-of-the-art performance in multiple disciplines within NLP. Fine-tuning the trained contextual models on task-specific datasets has been the key to achieving superior performance downstream. While fine-tuning these pre-trained models is straightforward for lexical applications (applications with only language modality), it is not trivial for multimodal language (a growing area in NLP focused on modeling face-to-face communication). Pre-trained models don't have the necessary components to accept two extra modalities of vision and acoustic. In this paper, we proposed an attachment to BERT and XLNet called Multimodal Adaptation Gate (MAG). MAG allows BERT and XLNet to accept multimodal nonverbal data during fine-tuning. It does so by generating a shift to internal representation of BERT and XLNet; a shift that is conditioned on the visual and acoustic modalities. In our experiments, we study the commonly used CMU-MOSI and CMU-MOSEI datasets for multimodal sentiment analysis. Fine-tuning MAG-BERT and MAG-XLNet significantly boosts the sentiment analysis performance over previous baselines as well as language-only fine-tuning of BERT and XLNet. On the CMU-MOSI dataset, MAG-XLNet achieves human-level multimodal sentiment analysis performance for the first time in the NLP community.~\footnote{This paper has been accepted in Association for Computational Linguistics (ACL) 2020 conference}
\end{abstract}

\blfootnote{* - Equal contribution}

\section{Introduction}
Human face-to-face communication flows as a seamless integration of language, acoustic, and vision modalities. In ordinary everyday interactions, we utilize all these modalities jointly to convey our intentions and emotions. Understanding this face-to-face communication falls within an increasingly growing NLP research area called multimodal language analysis \cite{zadeh2018proceedings}. The biggest challenge in this area is to efficiently model the three pillars of communication together. This gives artificial intelligence systems the capability to comprehend the multi-sensory information without disregarding nonverbal factors. In many applications such as dialogue systems and virtual reality, this capability is crucial to maintain the high quality of user interaction. 

The recent success of contextual word representations in NLP is largely credited to new Transformer-based \cite{vaswani2017attention} models such as BERT \cite{devlin2018bert} and XLNet \cite{yang2019xlnet}. These Transformer-based models have shown performance improvement across downstream tasks~\cite{devlin2018bert}. However, their true downstream potential comes from fine-tuning their pre-trained models for particular tasks \cite{devlin2018bert}. This is often done easily for lexical datasets which exhibit language modality only. However, this fine-tuning for multimodal language is neither trivial nor yet studied; simply because both BERT and XLNet only expect linguistic input. Therefore, in applying BERT and XLNet to multimodal language, one must either (a) forfeit the nonverbal information and fine-tune for language, or (b) simply extract word representations and proceed to use a state-of-the-art model for multimodal studies. 

In this paper, we present a successful framework for fine-tuning BERT and XLNet for multimodal input. Our framework allows the BERT and XLNet core structures to remain intact, and only attaches a carefully designed Multimodal Adaptation Gate (MAG) to the models. Using an attention conditioned on the nonverbal behaviors, MAG essentially maps the informative visual and acoustic factors to a vector with a trajectory and magnitude. During fine-tuning, this adaptation vector modifies the internal state of the BERT and XLNet, allowing the models to seamlessly adapt to the multimodal input. In our experiments we use the CMU-MOSI \cite{zadeh2016mosi} and CMU-MOSEI \cite{zadeh2018multimodal} datasets of multimodal language, with a specific focus on the core NLP task of multimodal sentiment analysis. We compare the performance of MAG-BERT and MAG-XLNet to the above (a) and (b) scenarios in both classification and regression sentiment analysis. Our findings demonstrate that fine-tuning these advanced pre-trained Transformers using MAG yields consistent improvement, even though BERT and XLNet were never trained on multimodal data. 

The contributions of this paper are therefore summarized as:
\begin{itemize}
    \item We propose an efficient framework for fine-tuning BERT and XLNet for multimodal language data. This framework uses a component called Multimodal Adaptation Gate (MAG) that introduces minimal overhead to both the models.
    \item MAG-BERT and MAG-XLNet set new state of the art in both CMU-MOSI and CMU-MOSEI datasets, when compared to scenarios (a) and (b). For CMU-MOSI, MAG-XLNet achieves performance on par with reported human performance. 
\end{itemize}

\section{Related Works}

The studies in this paper are related to the following research areas: 

\subsection{Multimodal Language Analyses} 
Multimodal language analyses is a recent research trend in natural language processing \cite{zadeh2018proceedings} that helps us understand language from the modalities of text, vision and acoustic. These analyses have particularly focused on the tasks of sentiment analysis \cite{poria2018multimodal}, emotion recognition \cite{zadeh2018multimodal}, and personality traits recognition \cite{park2014computational}. Works in this area often focus on novel multimodal neural architectures \cite{pham2018found,hazarika2018conversational} and multimodal fusion approaches \cite{liang2018multimodal,tsai2018learning}. 

Related to content in this paper, we discuss some of the models in this domain including TFN, MARN, MFN, RMFN and MulT. Tensor Fusion Network (TFN)~\cite{zadeh2017tensor} creates a multi-dimensional tensor to explicitly capture all possible interactions between the three modalities: unimodal, bimodal and trimodal. Multi-attention Recurrent Network (MARN)~\cite{zadeh2018multi} uses three separate hybrid LSTM memories that have the ability to propagate the cross-modal interactions. Memory Fusion Network~\cite{zadeh2018memory} synchronizes the information from three separate LSTMs through a multi-view gated memory. Recurrent Memory Fusion Network (RMFN)~\cite{liang2018multimodal} captures the nuanced interactions among the modalities in a multi-stage manner, giving each stage the ability to focus on a subset of signals. Multimodal Transformer for Unaligned Multimodal Language Sequences (MulT)~\cite{tsai2019multimodal} deploys three Transformers -- each for one modality -- to capture the interactions with the other two modalities in a self-attentive manner. The information from the three Transformers are aggregated through late-fusion.

\subsection{Pre-trained Language Representations }
Learning word representations from large corpora has been an active research area in NLP community \cite{mikolov2013distributed,pennington2014glove}. Glove~\cite{pennington2014glove} and Word2Vec~\cite{mikolov2013distributed} contributed to advancing the state-of-the-art of many NLP tasks. A major setback of these word representations is their non-contextual nature.
Recently, contextual language representation models trained on large text corpora have achieved state of the art results on several NLP tasks including question answering, sentiment classification,  part-of-speech (POS) tagging and similarity modeling\cite{peters2018deep,devlin2018bert}. The first two notable contextual representation based models were ELMO \cite{peters2018deep} and GPT \cite{radford2018improving}. However, they only captured unidirectional context and therefore, missed more nuanced interactions among words of a sentence.  BERT (Bidirectional  Encoder  Representations from Transformers) \cite{devlin2018bert}  outperforms both ELMO  and GPT since it can provide better representation through capturing bi-directional context using Transformers. XLNet\cite{dai2019transformer} gives new contextual representations through building an auto-regressive model capable of capturing all possible factorizations of the input. Fine-tuning pretrained models for BERT and XLNet has been a key factor in achieving state of the art performance for downstream tasks. Even though previous works have explored using BERT to model multimodal data \cite{sun2019videobert}, to the best of our knowledge, directly fine-tuning BERT or XLNet for multimodal data has not been explored in previous works.


\section{BERT and XLNet}
To better understand the proposed multimodal framework in this paper, we first present an overview of both the BERT and XLNet models. We start by quickly formalizing the operations within Transformer and Transformer-XL models, followed by an overview of BERT and XLNet.

\subsection{Transformer}

Transformer is a non-recurrent neural architecture designed for modeling sequential data \cite{vaswani2017attention}.
The superior performance of Transformer model is largely credited to a Multi-head Self-Attention module. Using this module, each element of a sequence is attended by conditioning on all the other sequence elements. Figure \ref{multi_modal_fig} summarizes internal operations of a Transformer layer (for $M$ such layers). Commonly, a Transformer uses an encoder-decoder paradigm. A stack of encoders is followed by a stack of decoders to map an input sequence to an output sequence. An additional embedding step with Positional Input Embedding is applied before the input goes through the stack of encoders and decoders. 


\subsection{Transformer-XL}
Transformer-XL~\cite{dai2019transformer} is an extension of the Transformer which offers two improvements: a) it enhances the capability of the Transformer to capture long-range dependencies (specifically for the case of context fragmentation), and b) it improves the capability to better predict first few symbols (which are often crucial for the rest of the sequence). It does so with a recurrence mechanism designed to pass context information from one segment to the next and a relative positional encoding mechanism to enable state reuse without causing temporal confusion.

\subsection{BERT}
BERT is a successful language model that provides rich contextual word representation \cite{devlin2018bert}. It follows an auto-encoding approach -- masking out a portion of input tokens and predicting those tokens based on all other non-masked tokens -- and thus learning a vector representation for the masked out tokens in that process.  
We use the variant of BERT used for \textit{Single Sentence Classification Tasks}. First, \textit{input embeddings} are generated from a sequence of word-piece tokens by adding token embeddings, segment embeddings and position embeddings . Then multiple Encoder layers are applied on top of these input embeddings. Each Encoder has a Multi-Head Attention layer and a Feed Forward layer, each followed by a residual connection with layer normalization. A special [CLS] token is appended in front of the input token sequence. So, for a $N$ length input sequence, we get $N+1$ vectors from the last Encoder layer -- the first of those vectors is used to predict the label of the input after that vector undergoes an affine transformation. 

\subsection{XLNet}
XLNet~\cite{yang2019xlnet} sets out to improve two critical aspects of the BERT model: a) independence among the masked out tokens and b) pretrain-finetune discrepancy in training vs inference, since inference inputs do not have masked out tokens. XLNet is an auto-regressive model and therefore, is free from the need of masking out certain tokens. However, auto-regressive models usually capture the unidirectional context (either forward or backward). XLNet can learn bidirectional context by maximizing likelihood over all possible permutations of factorization order. In essence, it randomly samples multiple factorization orders and trains the model on each of those orders. Therefore, it can model input by taking all possible permutations into consideration (in expectation).

XLNet utilizes two key ideas from Transformer-XL~\cite{dai2019transformer}: relative positioning and segment recurrence mechanism. Like BERT, it also has a Input Embedder followed by multiple Encoders. The Embedder converts the input tokens into vectors after adding token embedding, segment embedding and relative positional embedding information. Each encoder consists of a Multi-Head attention layer and a feed forward layer -- each followed by a residual addition and normalization layer. The embedder output is fed into the encoders to get a contextual representation of input. 


\begin{figure}[t!]
\includegraphics[width=1\linewidth]{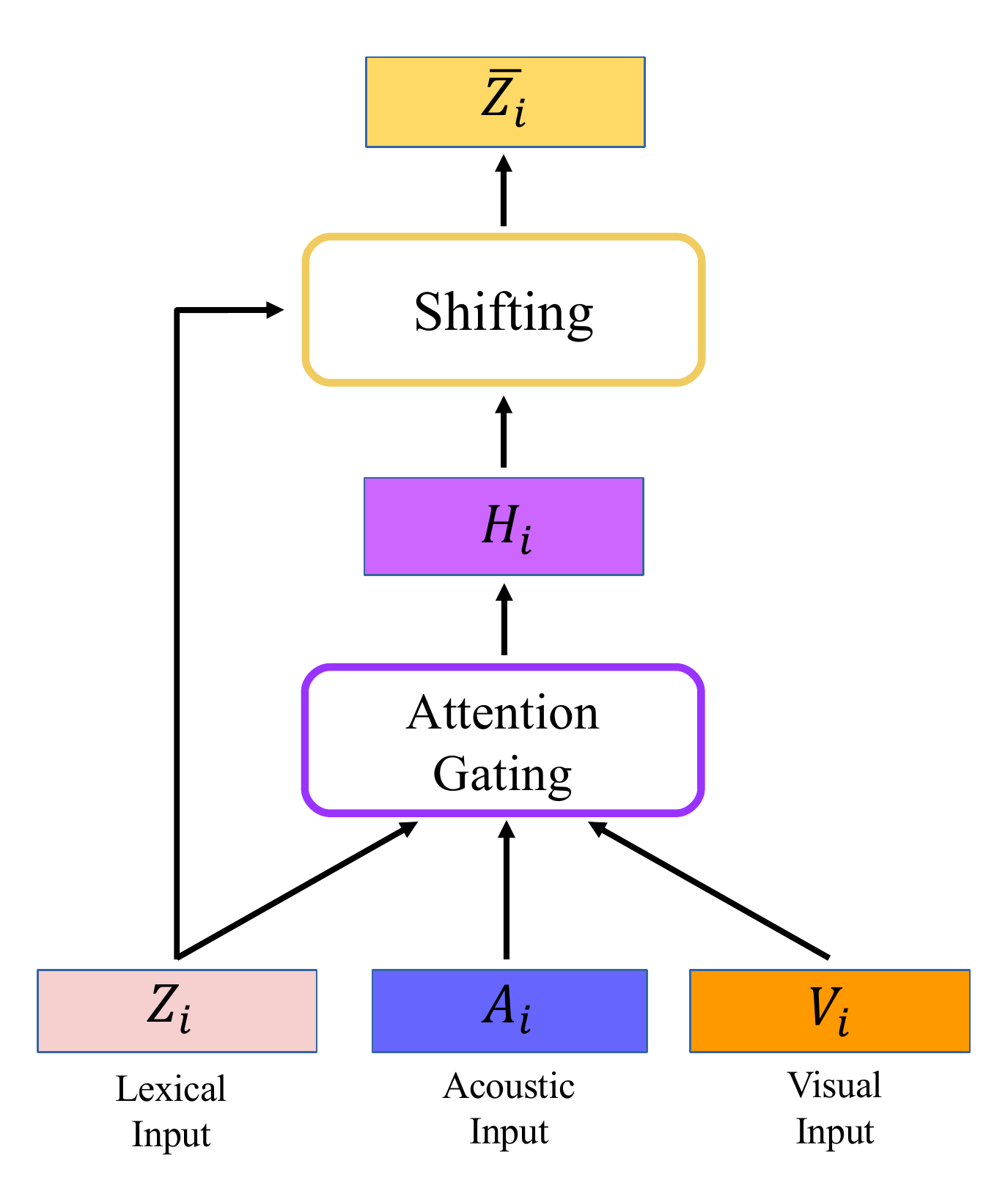}
\caption{Multimodal Adaptation Gate (MAG) takes as input a lexical input vector, as well as its visual and acoustic accompaniments. Subsequently, an attention over lexical and nonverbal dimensions is used to fuse the multimodal data into another vector, which is subsequently added to the input lexical vector (shifting). }
\label{fig_adaptation_gate}
\end{figure}

\section{Multimodal Adaptation Gate (MAG)}\label{model:multi_adaptation}
In multimodal language, a lexical input is accompanied by visual and acoustic information - simply gestures and prosody co-occurring with language. Consider a semantic space that captures latent concepts (positions in the latent space) for individual words. In absence of multimodal accompaniments, the semantic space is directly conditioned on the language manifold. Simply put, each word falls within some part of this semantic space, depending only on the meaning of the word in a linguistic structure (i.e. sentence). Nonverbal behaviors can have an impact on the meaning of words, and therefore on the position of words in this semantic space. Together, language and nonverbal accompaniments decide on the new position of the word in the semantic space. In this paper, we regard to this new position as addition of the language-only position with a displacement vector; a vector with trajectory and magnitude that shifts the language-only position of the word to the new position in light of nonverbal behaviors. This is the core philosophy behind the Multimodal Adaptation Gate (MAG). 

A particularly appealing implementation of such displacement is studied in RAVEN~\cite{wang2018words}, where displacements are calculated using cross-modal self-attention to highlight relevant nonverbal information. Figure \ref{fig_adaptation_gate} shows the studied MAG in this paper. Essentially, a MAG unit receives three inputs, one is purely lexical, one is visual, and the last one is acoustic. Let the triplet $(Z_i,A_i,V_i)$ denote these inputs for $i$th word in a sequence. We break this displacement into bimodal factors $[Z_i;A_i]$ and $[Z_i;V_i]$ by concatenating \textit{lexical vector} with acoustic and visual information respectively and use them to produce two gating vectors $g_{i}^{v}$ and  $g_{i}^{a}$:
\begin{equation}
    g_{i}^{v} = R(W_{gv}[Z_i;V_i] + b_v)\label{model_gate_vision}
\end{equation}
\begin{equation}
    g_{i}^{a} = R(W_{ga}[Z_i;A_i] + b_a)\label{model_gate_audio}
\end{equation}
where $W_{gv}$, $W_{ga}$ are weight matrices for visual and acoustic modality and $b_v$ and $b_a$ are scalar biases. $R(x)$ is a non-linear activation function. These gates highlight the relevant information in visual and acoustic modality conditioned on the lexical vector. 

We then create a non-verbal displacement vector $H_i$ by fusing together $A_i$ and $V_i$ multiplied by their respective gating vectors:
\begin{equation}
    H_i = g_{i}^{a} \cdot (W_a A_i) + g_{i}^{v} \cdot (W_v V_i) + b_H
\end{equation}
where $W_a$ and $W_v$ are weight matrices for acoustic and visual information respectively and $b_H$ is the bias vector.\newline
Subsequently, we use a weighted summation between $Z_i$ and its nonverbal displacement $H_i$ to create a \textit{multimodal vector} $\bar{Z_i}$:
\begin{equation}
    \bar{Z_i} = Z_i + \alpha H_i 
\end{equation}
\begin{equation}
    \alpha = min(\frac{\left\Vert Z_i \right \Vert _2}{\left\Vert H_i \right \Vert _2} \beta,1)
\end{equation}
where $\beta$ is a hyper-parameter selected through the cross-validation process. $\left\Vert Z_i \right \Vert _2$ and $\left\Vert H_i \right \Vert _2$ denote the $L_2$ norm of the $Z_i$ and $H_i$ vectors respectively. We use the scaling factor $\alpha$ so that the effect of non-verbal shift $H_i$ remains within a desirable range. Finally, we apply a layer normalization and dropout layer to $\bar{Z}_i$.

\begin{figure}[t!]
\includegraphics[width=1\linewidth,height=\textwidth]{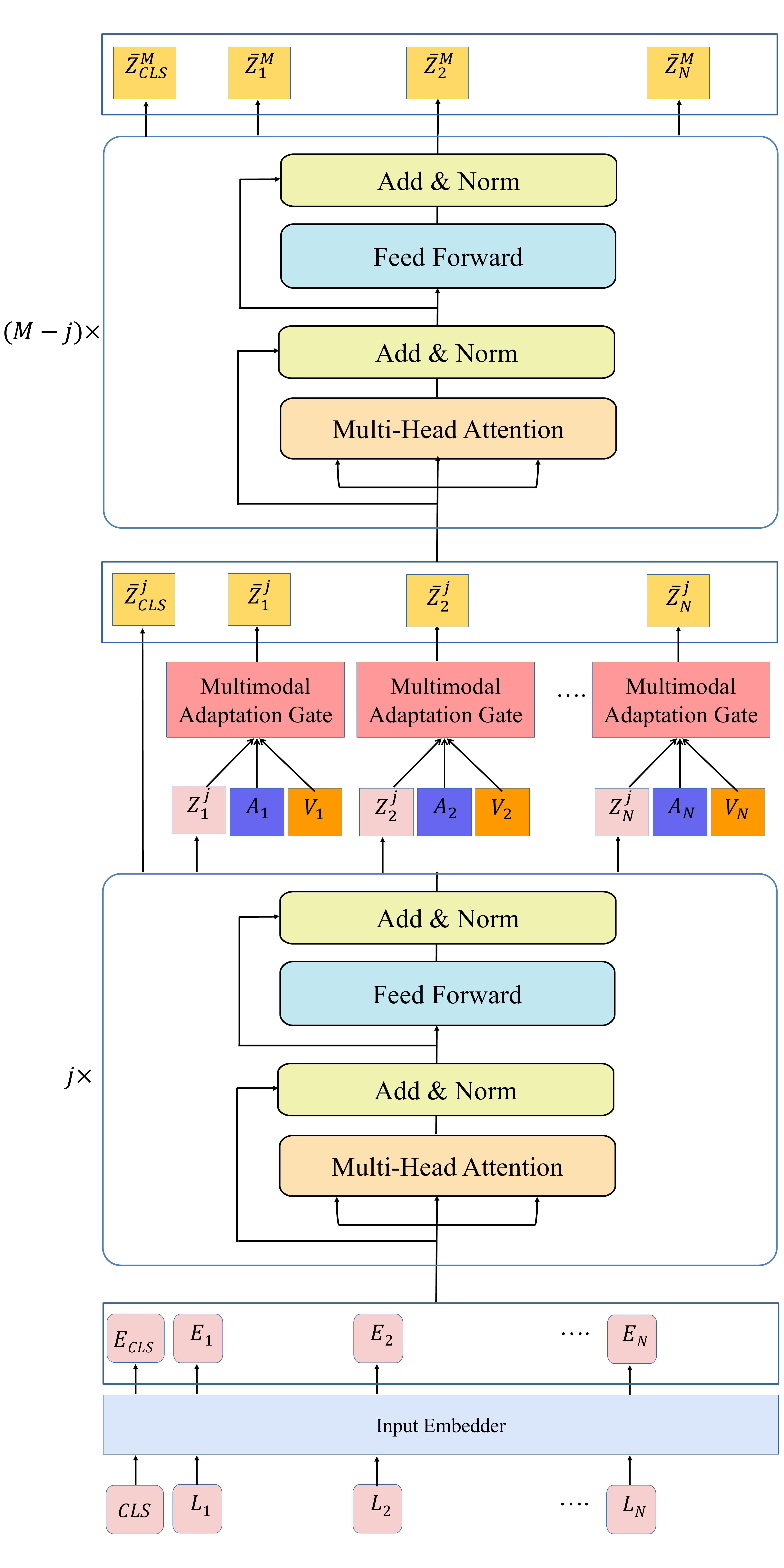}
\caption{Best viewed zoomed in and in color. The Transformer architecture of BERT/XLNet with MAG applied at $j$th layer. We consider a total of $M$ layers within the pretrained Transformer. MAG can be applied at different layers of the pretrained Transformers.  }
\label{multi_modal_fig}
\end{figure}

\subsection{MAG-BERT}
MAG-BERT is a combination of MAG applied to a certain layer of BERT network (Figure \ref{multi_modal_fig} demonstrates the structure of MAG-BERT as well as MAG-XLNet). Essentially, at each layer, BERT contains lexical vectors for $i$th word in the sequence. For the same word, nonverbal accompaniments are also available in multimodal language setup. MAG essentially forms an attachment to the desired layer in BERT; an attachment that allows for multimodal information to leak into the BERT model and displace the lexical vectors. The operations within MAG allows for the lexical vectors within BERT to adapt to multimodal information by changing their positions within the semantic space. Aside from the attachment of MAG, no change is made to the BERT structure. 

Given an $N$ length language sequence $L = [L_1, L_2,\dots L_N]$ carrying word-piece tokens, a [CLS] token is appended to $L$ so that we can use it later for class label prediction. Then, we input $L$ to the \textbf{Input Embedder} which outputs $E = [E_{CLS}, E_1,E_2, \dots E_N]$ after adding token, segment and position embeddings. Then, we input $E$ to the first Encoding layer and then apply $j$ Encoders on it successively. After that encoding process, we get the output $Z^j = [Z^j_{CLS},Z^j_1,Z^j_2,\dots Z^j_N]$ which denotes the Lexical Embeddings after $j$ layers of Encoding.

For injecting audio-visual information into these embeddings, we  prepare a sequence of triplets $[(Z_i^j,A_i,V_i):\forall i \in \{CLS,[1,N]\}]$ by pairing $Z_i^j$ with the corresponding $(A_i,V_i)$. Each of these triplets are passed through the \textbf{Multimodal Adaptation Gate} which transforms the $i$th triplet into $\bar{Z}_{i}^j$ -- a unified multimodal representation of the corresponding Lexical Embedding.

As there exists $M=12$ Encoder layers in our BERT model, we input $\bar{Z^j} = [\bar{Z}_{1}^j,\bar{Z}_{2}^j,\dots \bar{Z}_{N}^j]$ to the next Encoder and apply $M-j$ Encoder layers on it successively. At the end, we get $\bar{Z}^M$ from the $M$th Encoder layer. As the first element $\bar{Z}_{CLS}^M$ represents the [CLS] token, it has the information necessary to make a class label prediction. Therefore, $\bar{Z}_{CLS}^M$ goes through an affine transformation to produce a single real-value which can be used to predict a class label. 

\subsection{MAG-XLNet}
Like MAG-BERT, MAG-XLNet also has the capability of injecting audio-visual information at any of its layers using MAG. At each position $j$ of any of its layer, it holds the lexical vector corresponding to that position. Utilizing the audio-visual information available for that position, it can invoke MAG to get an appropriately shifted lexical vector in multimodal space. Although it mostly follows the general paradigm presented in Figure \ref{multi_modal_fig} verbatim, it uses the XLNet specific Embedder and Encoders. One other key difference is the position of the [CLS] token. Unlike BERT, the [CLS] token is appended at the right end of the input token sequence, and therefore in all the intermediate representations, the vector corresponding to the [CLS] will be the rightmost one. Following the same logic, the output from the final Encoding layer will be $\bar{Z}^M = [\bar{Z}_{1}^M,\bar{Z}_{2}^M,\dots \bar{Z}_{N}^M, \bar{Z}_{CLS}^M]$. The last item, $\bar{Z}_{CLS}^M$ can be used for class label prediction after it goes through an affine transformation.

\section{Experiments}
In this section we outline the experiments in this paper. We first start by describing the datasets, followed by description of extracted features, baselines, and experimental setup. 

\subsection{CMU-MOSI Dataset}
CMU-MOSI (CMU Multimodal Opinion Sentiment Intensity) is a dataset of multimodal language specifically focused on multimodal sentiment analysis~\cite{zadeh2016mosi}. CMU-MOSI contains 2199 video segments taken from 93 Youtube movie review videos. The dataset has real-valued high-agreement sentiment intensity annotations in the range $[-3,+3]$.


\subsection{Computational Descriptors}
For each modality, the following computational descriptors are available: 

\noindent \textbf{Language:} 
We transcribe the videos using Youtube API followed by manual correction.


\noindent \textbf{Acoustic:} 
COVAREP~\cite{degottex2014covarep} is used to extract the following relevant features: fundamental frequency, quasi open quotient, normalized amplitude quotient, glottal source parameters (H1H2, Rd, Rd conf), VUV, MDQ, the first 3 formants, PSP, HMPDM 0-24 and HMPDD 0-12, spectral tilt/slope of wavelet responses (peak/slope), MCEP 0-24.  


\noindent \textbf{Visual:} For the visual modality, the Facet library \cite{facet} is used to extract a set of visual features including facial action units, facial landmarks, head pose, gaze tracking and HOG features. 

For each word, we align  all three modalities following the convention established in ~\cite{chen2017multimodal}. Firstly, the word alignment between language and audio is obtained using forced alignment ~\cite{yuan2008speaker}. Afterwards, the boundary of each word denotes the co-occurring visual and acoustic features (FACET and COVAREP). Subsequently, for each word, the co-occurring acoustic and visual features are averaged across each feature -- thus achieving  $A_i$ and $V_i$  vectors corresponding to word $i$. 


\subsection{Baseline Models }
\label{sec:baseline_models}
We compare the performance of MAG-BERT and MAG-XLNet to a variety of state-of-the-art models for multimodal language analysis. These models are trained using extracted BERT and XLNet word embeddings as their language input:

\noindent \textbf{TFN (Tensor Fusion Network)} explicitly models both intra-modality and inter-modality dynamics \cite{zadeh2017tensor} by creating a multi-dimensional tensor that captures unimodal, bimodal and trimodal interactions across three modalities. 

\noindent \textbf{MARN (Multi-attention Recurrent Network)} models view-specific interactions using hybrid LSTM memories and cross-modal
interactions using a Multi-Attention Block (MAB)~\cite{zadeh2018multi}. 

\noindent \textbf{MFN (Memory Fusion Network)} has three separate LSTMs to model each modality separately and a multi-view gated memory to synchronize among them~\cite{zadeh2018memory}.


\noindent \textbf{RMFN (Recurrent Memory Fusion Network)} captures intra-modal and inter-modal information through recurrent multi-stage fashion~\cite{liang2018multimodal}. 

\noindent \textbf{MulT (Multimodal Transformer for Unaligned Multimodal Language Sequence)} uses three sets of Transformers and combines their output in a late fusion manner to model a multimodal sequence~\cite{tsai2019multimodal}. We use the aligned variant of the originally proposed model, which achieves superior performance over the unaligned variant. 

We also compare our model to fine-tuned \textbf{BERT} and \textbf{XLNet} using language modality only to measure the success of the MAG framework. 

\subsection{Experimental Design}

All the models in this paper are trained using Adam \cite{kingma2014adam} optimizer with learning rates between $\{0.001,0.0001,0.00001\}$. We use dropouts of $\{0.1,0.2,0.3,0.4,0.5\}$ for training each model. LSTMs in TFN, MARN, MFN, RMFN, LFN use latent size of $\{16,32,64,128\}$. For MulT, we use $\{3,5,7\}$ layers in the network and $\{1,3,5\}$ attention heads. All models use the designated validation set of CMU-MOSI for finding best hyper-parameters.

We perform two different evaluation tasks on CMU-MOSI datset: i)  Binary Classification, and ii) Regression. We formulate it as a regression problem and report Mean-absolute Error (MAE) and the correlation of model predictions with true labels. Besides, we convert the regression outputs into categorical values to obtain binary classification accuracy (BA) and F1 score. Higher value means better performance for all the metrics except MAE. We use two evaluation metrics for BA and F1, one used in \cite{zadeh2018multimodal} and one used in \cite{tsai2019multimodal}. 
\section{Results and Discussion}

\begin{table}[t!]
\begin{center}
\small
\fontsize{9}{10}\selectfont
\setlength{\tabcolsep}{4.5pt}
\renewcommand{\arraystretch}{1.2}
\begin{tabular}{l|cc|cc}
Task Metric & BA$\uparrow$ & F1$\uparrow$ & MAE$\downarrow$ & Corr$\uparrow$ \\
\hline
\multicolumn{5}{|c|}{Original (glove)} \\
\hline 
TFN & 73.9/-- & 73.4/--  & 0.970/-- & 0.633/-- \\
MARN & 77.1/-- & 77.0/--  & 0.968/-- & 0.625/-- \\
MFN & 77.4/-- & 77.3/--  & 0.965/-- & 0.632/-- \\
RMFN & 78.4/-- & 78.0/--  & 0.922/-- & 0.681/-- \\
LFN & 76.4/-- & 75.7/--  & 0.912/-- & 0.668/-- \\
MulT &  --/83.0 & --/82.8 & --/0.871 & --/0.698 \\
\hline 
\hline 
\multicolumn{5}{|c|}{BERT} \\
\hline 
TFN & 74.8/76.0 & 74.1/75.2  & 0.955 & 0.649 \\
MARN & 77.7/78.9 & 77.9/78.2  & 0.938 & 0.691 \\
MFN & 78.2/79.3 & 78.1/78.4  & 0.911 & 0.699 \\
RMFN & 79.6/80.7 & 78.9/79.1  & 0.878 & 0.712 \\
LFN & 79.1/80.2 & 77.3/78.1  & 0.899 & 0.701 \\
MulT &  81.5/84.1 & 80.6/83.9 & 0.861 & 0.711 \\
\hline
BERT& 83.5/85.2 & 83.4/85.2  & 0.739 & 0.782\\
MAG-BERT & \textbf{84.2}/\textbf{86.1} & \textbf{84.1}/\textbf{86.0} & \textbf{0.712} & \textbf{0.796} \\
\hline
\hline 
\multicolumn{5}{|c|}{XLNet} \\
\hline 
TFN & 78.2/80.1 & 78.2/78.8  & 0.914 & 0.713 \\
MARN & 78.3/79.5 & 78.8/79.6  & 0.921 & 0.707 \\
MFN & 78.3/79.9 & 78.4/79.1  & 0.898 & 0.713 \\
RMFN & 79.1/81.0 & 78.6/80.0  & 0.901 & 0.703 \\
LFN & 80.2/82.9 & 79.1/81.6  & 0.862 & 0.701 \\
MulT &  81.7/84.4 & 80.4/83.1 & 0.849 & 0.738 \\
\hline
XLNet& 84.7/86.7 & 84.6/86.7  & 0.676 & 0.812\\
MAG-XLNet & \textbf{85.7}/\textbf{87.9} & \textbf{85.6}/\textbf{87.9} & \textbf{0.675} & \textbf{0.821} \\
\hline
Human & 85.7/- & 87.5/- & 0.710 & 0.820 \\
\hline
\hline
\end{tabular}
\end{center}
\caption{\label{table:mosi_performance}Sentiment prediction results on CMU-MOSI dataset. Best results are highlighted in bold. MAG-BERT and MAG-XLNet achieve superior performance than the baselines and their language-only finetuned counterpart. BA denotes binary accuracy (higher is better, same for F1), MAE denotes Mean-absolute Error (lower is better), and Corr is Pearson Correlation (higher is better). For BA and F1, we report two numbers: the number on the left side of ``/'' is measures calculated based on \cite{zadeh2018multi} and the right side is measures calculated based on \cite{tsai2019multimodal}. Human performance for CMU-MOSI is reported as \cite{zadeh2018memory}. }

\end{table}
\begin{table}[t!]
\fontsize{9}{10}\selectfont
\renewcommand{\arraystretch}{1.2}
\setlength\tabcolsep{1.2pt}
\begin{tabular}{l|c|c|c|c|c|c|c|c|c}

                               Model & E    & 1    & 4    & 6    & 8    & 12 & A & $\bigoplus$ & $\bigodot$  \\ \hline
\multicolumn{1}{l|}{MAG-XLNet} & 80.1 & 85.6 & 84.1 & 84.1 & 83.8 & 83.6  & 64.0 & 60.0 & 55.8\\
\hline
\end{tabular}
\caption{\label{table:mag_layer}Results of variations of XLNet model: MAG applied at different layers of the XLNet model, input-level concatenation and addition of all modalities. ``E'' denotes application of MAG immediately after embedding layer of the XLNet and ``A'' denotes applying MAG after the embedding layer and all the subsequent Encoding layers. $\bigoplus$ and $\bigodot$ denote input-level addition and concatenation of all modalities respectively. MAG applied at initial layers performs better overall. }

\end{table}

\newcolumntype{C}[1]{>{\centering}m{#1}}
\begin{table*}[t!]
  \begin{center}
  \small
  \begin{tabular}{C{0.2cm} >{\small}m{10.0cm} | C{1.2cm} | C{1.2cm} | C{1.2cm}  }
    \toprule
    \# &\centering {Spoken words + \\ acoustic and visual behaviors}   &Ground Truth & MAG-XLNet & XLNet\tabularnewline \midrule
    
1& \colorbox{red!50}{“And it really just lacked what made the other movies more enjoyable.”} + \colorbox{red}{Frustrated and disappointed tone} & \colorbox{red}{-1.4}& \colorbox{red}{-1.41}& \colorbox{red!50}{-0.9} \tabularnewline \midrule
2& \colorbox{green!20}{“But umm I liked it.”} + \colorbox{green}{Emphasis on tone} + \colorbox{green}{positive shock through sudden eyebrow raise}  & \colorbox{green}{1.8}& \colorbox{green}{1.9}& \colorbox{green!20}{1.2} \tabularnewline \midrule
3& \colorbox{green!20}{“Except their eyes are kind of like this welcome to the polar express.”} + \colorbox{red}{tense voice} + \colorbox{red}{frown expression} & \colorbox{red!50}{-0.6}& \colorbox{red!50}{-0.6}& \colorbox{green!20}{0.8} \tabularnewline \midrule
4& \colorbox{black!30}{“Straight away miley cyrus acting miley cyrus, or lack of, she had this} \colorbox{black!30}{same expression throughout the entire film”} + \colorbox{red}{sarcastic voice} + \colorbox{red}{frustrated facial expression} & \colorbox{red}{-1.0} & \colorbox{red}{-1.2} & \colorbox{black!30}{0.2} \tabularnewline \midrule

  \end{tabular}
  \captionof{table}{Examples from the CMU-MOSI dataset. The ground truth sentiment labels are between strongly negative (-3) and strongly positive (+3). For each example, we show the Ground Truth and prediction output of both the MAG-XLNet and XLNet. XLNet seems to be replicating language modality mostly while MAG-XLNet is integrating the non-verbal information successfully.
  }  \label{table:visualization}
  \end{center}
\end{table*}




Table \ref{table:mosi_performance} shows the results of the experiments in this paper. We summarize the observations from the results in this table as following: 

\subsection{Performance of MAG-BERT} In all the metrics across the CMU-MOSI dataset, we observe that performance of MAG-BERT is superior to state-of-the-art multimodal models that use BERT word embeddings. Furthermore, MAG-BERT also performs superior to fine-tuned BERT. This essentially shows that the MAG component is allowing the BERT model to adapt to multimodal information during fine-tuning, thus achieving superior performance. 

\subsection{Performance of MAG-XLNet} A similar performance trend to MAG-BERT is also observed for MAG-XLNet. Besides superior performance than baselines and fine-tuned XLNet, MAG-XLNet achieves near-human level performance for CMU-MOSI dataset. Furthermore, we train MulT using the fine-tuned XLNet embeddings and get the following performance: $ 83.6/85.3 , 82.6/84.2 , 0.810 , 0.759 $ which is lower than both MAG-XLNet and XLNet. It is notable that the p-value for student t-test between MAG-XLNet and XLNet in Table~\ref{table:mosi_performance} is lower than $10e-5$ for all the metrics. 

 The motivation behind the experiments reported in Table \ref{table:mosi_performance} is as follows: we extracted word embeddings from pre-trained BERT and XLNet models and trained the baseline models using those embeddings. Since BERT and XLNet are often perceived to provide better word embeddings than Glove, it is not fair to compare MAG-BERT/MAG-XLNet with previous models trained with Glove embeddings. Therefore, we retrain previous works using BERT/XLNet embeddings to establish a more fair comparison between proposed approach in this paper, and previous work. Based on the information from Table \ref{table:mosi_performance}, we observe that MAG-BERT/MAG-XLNet models outperforms various baseline models using BERT/XLNet/Glove models substantially.

\subsection{Adaptation at Different Layers} We also study the effect of applying MAG at different encoder layers of the  XLNet. Specifically, we first apply the MAG to the output of the embedding layer. Subsequently, we apply the MAG to the layer $j \in \{1,4,6,8,12\}$ of the XLNet. Then, we apply MAG at all the XLNet layers. From Table~\ref{table:mag_layer}, we observe that earlier layers are more suitable for application of MAG. 

We believe that earlier layers allow for better integration of the multimodal information, as they allow the word shifting to happen from the beginning of the network. If the semantics of words should change based on the nonverbal accompaniments, then initial layers should reflect the semantic shift, otherwise, those layers are only working unimodally. Besides, the higher layers of BERT learn more abstract and higher-level information about the syntactic and semantic structure of linguistic features~\cite{coenen2019visualizing}. Since, the acoustic and visual information present in our model corresponds to each word in the utterance, it will be more difficult for the MAG to shift the vector extracted from a later layer since that vector's information will be very abstract in nature.

\subsection{Input-level Concatenation and Addition}
From Table \ref{table:mag_layer}, we see that both input-level concatenation and addition of modalities perform poorly. For Concatenation, we simply concatenate all the modalities. For Addition, we add the audio and visual information to the language embedding after mapping both of them to the language dimension. These results demonstrate the rationale behind using an advanced fusion mechanism like MAG.


\subsection{Results on Comparable Datasets} We also perform experiments on the CMU-MOSEI dataset \cite{zadeh2018multimodal} to study the generalization of our approach to other multimodal language datasets. Unlike CMU-MOSI which has sentiment annotations at utterance level, CMU-MOSEI has sentiment annotations at sentence level. The experimental methodology for CMU-MOSEI is similar to the original paper. For the sake of comparison, we suffice\footnote{Since Transformer based models take a long time to train for CMU-MOSEI} to comparing the binary accuracy and f1 score for the top 3 models in Table \ref{table:mosi_performance}. In BERT category, we compare the performance of MulT (with BERT embeddings), BERT and MAG-BERT which are respectively as follows: $[83.5, 82.9]$ for MulT, $[83.9, 83.9]$ for BERT, and $[84.7, 84.5]$ for MAG-BERT. Similarly for XLNET category, the results for MulT (with XLNet embeddings), XLNet and MAG-XLNet are as follows: $[84.1, 83.7]$ for MulT, $[85.4,85.2]$ for XLNet and $[85.6,85.7]$ for MAG-XLNet. Therefore, superior performance of MAG-BERT and MAG-XLNet also generalizes to CMU-MOSEI dataset. 

\subsection{Fine-tuning Effect} We study whether or not the superior performance of the MAG-BERT and MAG-XLNet is related to successful finetuning of the models, or related to other factors e.g. any transformer with architecture like BERT or XLNet would achieve superior performance regardless of being pretrained. By randomly initializing the weights of BERT and XLNet within MAG-BERT and MAG-XLNet, we get the following performance on BA for the CMU-MOSI: 70.1 and 70.7 respectively. This indicates that the success of the MAG-BERT and MAG-XLNet is due to successful fine-tuning. Even on the larger CMU-MOSEI dataset we get BA of 76.8 and 78.4 for MAG-BERT and MAG-XLNet, which further substantiates the fact that fine-tuning is successful using MAG framework.

\subsection{Qualitative Analysis}
In Table~\ref{table:visualization}, we present some examples where MAG-XLNet adjusted sentiment intensity properly by taking into account nonverbal information. The examples demonstrate that MAG-XLNET can successfully integrate the non-verbal modalities with textual information.  

In both Example-1 and Example-2, XLNet correctly predicted the polarity of the displayed emotion. However, additional information was present in the acoustic and visual domain which XLNet could not utlize. Given those information, MAG-XLNet could better predict the magnitude of emotion displayed in both cases.

Although the emotion in the text of Example-3 can be portrayed as a bit positive, the tense voice and frown expression helps MAG-XLnet reverse the polarity of predicted emotion. Similarly, the text in Example-4 is mostly neutral, but MAG-XLNet can predict the negative emotion through the sarcastic vocal and frustrated facial expression.

\section{Conclusion}

In this paper, we introduced a method for efficiently finetuning large pre-trained Transformer models for multimodal language. Using a proposed Multimodal Adaptation Gate (MAG), BERT and XLNet were successfully fine-tuned in presence of vision and acoustic modalities. MAG essentially poses the nonverbal behavior as a vector with a trajectory and magnitude,  which is subsequently used to shift lexical representations within the pre-trained Transformer model. A unique characteristic of MAG is that it makes no change to the original structure of BERT or XLNet, but rather comes as an attachment to both models. Our experiments demonstrated the superior performance of MAG-BERT and MAG-XLNet. The code for both MAG-BERT and MAG-XLNet are publicly available here~\footnote{\giturl}



\section*{Acknowledgement}
 This research was supported in part by grant W911NF-15-1-0542 and W911NF-19-1-0029 with the US Defense Advanced Research Projects Agency (DARPA) and the Army Research Office (ARO). Authors AZ and LM were supported by the National Science Foundation (Awards \#1750439 \#1722822) and National Institutes of Health. Any opinions, findings, and conclusions or recommendations expressed in this material are those of the author(s) and do not necessarily reflect the views of US Defense Advanced Research Projects Agency, Army Research Office, National Science Foundation or National Institutes of Health, and no official endorsement should be inferred.

\bibliography{ms}
\bibliographystyle{ms}

\end{document}